\documentclass[conference]{IEEEtran}
\ifCLASSINFOpdf
   \usepackage{graphicx}
   \usepackage{epstopdf} 

    \usepackage{amsmath,amssymb}
    \usepackage{graphicx}
    \usepackage[colorlinks=true,citecolor=black,linkcolor=black,urlcolor=blue,bookmarks=false]{hyperref}
    \usepackage{tikz,graphics,color,fullpage,float,epsf,caption,subcaption}
\else
\fi

\begin{document}
\begin{NoHyper}
%
\title{EBBIOT: A Low-complexity Tracking Algorithm for Surveillance in IoVT using Stationary Neuromorphic Vision Sensors}



%

\author{
  \IEEEauthorblockN{%
    Jyotibdha Acharya\IEEEauthorrefmark{1}\,
    Andres Ussa Caycedo\IEEEauthorrefmark{2}\,
    Vandana Reddy Padala\IEEEauthorrefmark{1}\,
    Rishi Raj Singh Sidhu\IEEEauthorrefmark{1}\\
    Garrick Orchard\IEEEauthorrefmark{2}\,
    Bharath Ramesh\IEEEauthorrefmark{2}\,
    Arindam Basu\IEEEauthorrefmark{1}
  }
  \IEEEauthorblockA{%
    \IEEEauthorrefmark{1}School of Electrical and Electronic Engineering, Nanyang Technological University, Singapore \\
    \IEEEauthorrefmark{2}The N.1 Institute for Health, National University of Singapore, Singapore \\
    jyotibdh001@e.ntu.edu.sg, lsiacuc@nus.edu.sg, bharath.ramesh03@u.nus.edu, arindam.basu@ntu.edu.sg, \\\{vandanapadala27 $|$ rishirajsidhu $|$ garrickorchard\}@gmail.com \\
  }
}


\maketitle

\begin{abstract}
In this paper, we present EBBIOT--a novel paradigm for object tracking using stationary neuromorphic vision sensors in low-power sensor nodes for the Internet of Video Things (IoVT). Different from fully event based tracking or fully frame based approaches, we propose a mixed approach where we create event-based binary images (EBBI) that can use memory efficient noise filtering algorithms. We exploit the motion triggering aspect of neuromorphic sensors to generate region proposals based on event density counts with $> 1000X$ less memory and computes compared to frame based approaches. We also propose a simple overlap based tracker (OT) with prediction based handling of occlusion. Our overall approach requires $7X$ less memory and $3X$ less computations than conventional noise filtering and event based mean shift (EBMS) tracking. Finally, we show that our approach results in significantly higher precision and recall compared to EBMS approach as well as Kalman Filter tracker when evaluated over $1.1$ hours of traffic recordings at two different locations.
\end{abstract}

\IEEEpeerreviewmaketitle
\belowdisplayskip=1pt
\abovedisplayskip=1pt
\section{Introduction}
\label{sec:intro}
Internet of Things (IoT) is a rapidly growing phenomenon where millions of connected sensors are distributed to improve a variety of applications ranging from precision agriculture to smart factories. Among the sensors, cameras offer unique opportunities due to the wealth of information they provide \cite{iovt_2030} at the cost of hugely increased bandwidth and energy to wirelessly transmit the huge volume of video data. The unique challenges and opportunities offered by camera sensors has led to a sub-field of IoT called the internet of video things (IoVT). Edge computing becomes important in this case to process data locally to reduce wireless transmission \cite{jetcas_review}. Neuromorphic sensors and processors offer an unique low power solution for this case.



In the past, neuromorphic vision sensors (NVS) have been employed for a variety of applications and tasks including microsphere tracking, multiple person tracking, vehicle-speed estimation, controlling robotic-arm, gesture recognition, etc \cite{Syst2012}\cite{Delbruck2013}\cite{Amir}. While the role of these sensors in IoVT have been envisioned \cite{jetcas_review}, there has not been any concrete work showing details of resource (energy, area) required by NVS based solutions for IoVT. 


Object tracking forms the essential first step in most computer vision applications. Research work on tracking using NVS  has  mostly been focused on taking advantage of the high temporal  resolution  to  faithfully  track  high  speed objects which is a problem for frame based cameras \cite{Syst2012}\cite{Delbruck2013}. 
Mean shift \cite{Delbruck2013}, combination of CNN and particle filtering \cite{Liu2016} and Kalman Filters \cite{Camunas-mesa2018} have been employed in the past for tracking NVS outputs. While such applications demonstrate the ability of NVS based systems to handle complex tasks, they do not show their applicability to resource constrained systems which is a hallmark of IoT.

In this paper, we propose EBBIOT--a low-complexity tracking algorithm for surveillance applications in IoVT using a NVS. The focus of our approach is to make the whole system less memory intensive (thus reducing chip area) and less computationally complex leading to savings in energy. Different from purely event based or purely frame based approaches, we accumulate the events from a NVS into a binary image and perform tracking on these frames. We further propose an event density based region proposal network (RPN) that has far less computations compared to traditional RPN. Lastly, we demonstrate a simple overlap based tracker (OT) that requires far less resources than the conventional event based mean shift (EBMS) \cite{Delbruck2013} while producing superior performance compared to it. This is of immense importance when using NVS for IoVT applications like remote surveillance where long battery life of the sensor node is critical. We describe the details of our approach in the following sections.

\section{Materials and Methods}
\label{sec:materials}
\begin{figure}
\hspace{-0.2cm}
    \includegraphics[width=0.5\textwidth]{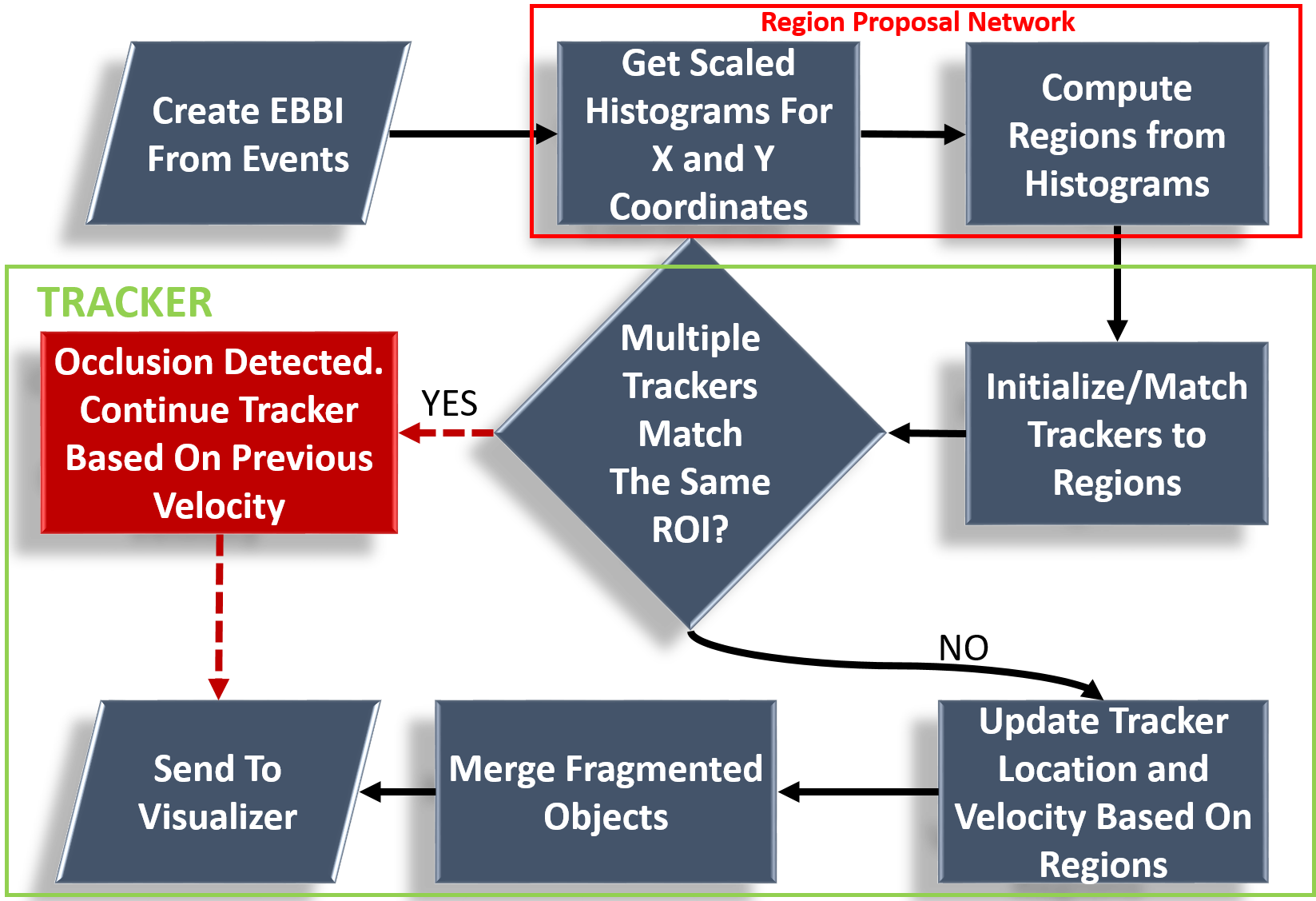}
    \caption{Flowchart depicting all the important blocks in the system: binary frame generation, region proposal and overlap based tracking.}
    \label{fig:flowchart}
    \vspace{-0.5cm}
\end{figure}
NVS differ from a traditional frame based sensor in that they output an event $[e_i = (x_i, y_i, t_i, p_i)]$ whenever there is a positive or negative change in the intensity of light falling on that pixel. Here, $e_i$ represents the event, $(x_i,y_i)$ represent the event’s location on the sensor array, $t_i$ represents the time stamp of the event generally at a microsecond resolution and $p_i$  = 1 (ON Event) whenever the change in intensity is increases beyond a threshold and $p_i$ = -1 (OFF Event) whenever the change in intensity decreases below a threshold \cite{davis}. We define a list of symbols that we will use throughout the paper:

\begin{tabular}{cp{0.6\textwidth}}
  $A\times B$ & Image resolution \\
  $B_t$ & Number of bits to store time stamp $t_i$ \\
  $N_T$ & Number of trackers\\
  $t_F$ & Frame duration\\
  $p$ & Neighbourhood size for noise filtering
\end{tabular}\\

The sensor used in this work is DAVIS \cite{davis} with $A=240$ and $B=180$. Also, we use $t_F=66$ ms which is sufficient for tracking vehicles--a longer exposure is needed for humans. A flowchart depicting the entire algorithm pipeline is shown in Figure \ref{fig:flowchart} and is described in details in the following subsections.

\subsection{Event based Binary Image and Noise Filtering}
\label{sec:ebbi}
\begin{figure}
    \centering
    \includegraphics[width=0.50\textwidth]{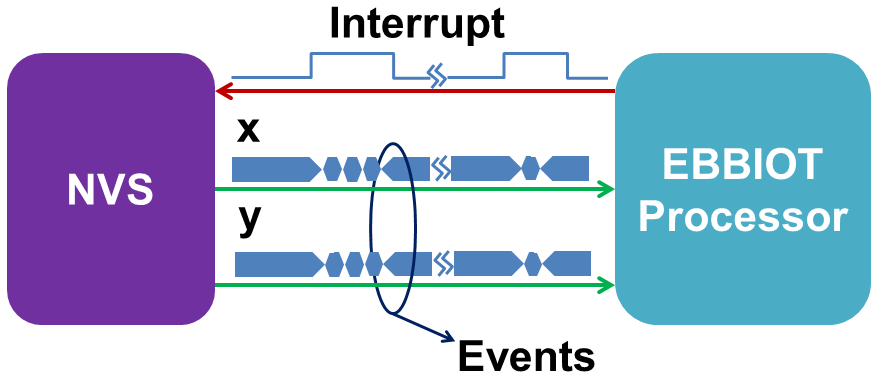}
    \caption{Timing diagram showing interrupt driven operation of the NVS for duty cycled low power operation.}
    \label{fig:timing}
    \vspace{-0.5cm}
\end{figure}

While most work on NVS has focussed on its event driven nature where number of computations are proportional to event rate, noise prevalent in such sensors invariably lead to spurious spikes even in the absence of any objects in the scene \cite{Padala2018}. This is a problem for IoT nodes which rely on saving energy by heavy duty cycling--using the NVS events as interrupts would rarely allow the processor to sleep. 

Instead, we propose to use an interrupt based sensing scheme where the EBBIOT processor generates an interrupt at regular time intervals $t_F$ to collect the events accumulated since the last interrupt (Figure \ref{fig:timing}). Such a scheme makes it possible to interface NVS with others FPGA and microprocessors commonly used in IoT. This scheme is feasible for two reasons:
\begin{itemize}
    \item Frame rates  ($\approx 15$ Hz)  are  good  enough  for  traffic  surveillance  as shown later in the paper. This scheme loses appeal as $t_F$ becomes smaller.
    \item We exploit the fact that the pixels firing an event are not reset till the event is readout in an NVS. Thus it can effectively  store a  binary image of events occurring while the processor is sleeping. In other words, we reuse the sensor as a memory.
    \end{itemize}
Since we readout a binary image with only one possible event per pixel (ignoring polarity), we call the image an event based binary image (EBBI). Note that the NVS is always awake in this scenario and it is the processor which goes to sleep and wakes up regularly. However, this binary image is useful only if sufficient information can be extracted from it--we show corresponding results in later sections. For a binary frame, noise removal may be easily done by a median filter \cite{Gonzalez:2006:DIP:1076432} (with patch size $p\times p$) since spurious events result in salt and pepper noise. In this work, we used $p=3$. The total number of computes per pixel of the filtered image is then equal to incrementing a counter every time a $1$ is encountered in the $p^2$ pixels of that patch followed by a comparison with $\left\lfloor p^2/2\right\rfloor$. This has to be added with the memory writes for creating the EBBI (ignoring memory reads due to lower energy requirement). The total memory requirement is twice the frame size--one frame to store the original image and one for the filtered version. We chose to keep the original frame since it might carry more information necessary for classification at a later stage. Thus we can summarize the computation $C_{EBBI}$ and memory $M_{EBBI}$ required by the proposed method as:
\begin{align}
\label{eq:memory_EBBI}
    C_{EBBI} &\approx (\alpha p^2+2)\times A\times B (\because p<<A,B)\notag\\
    M_{EBBI} &= 2\times A\times B
\end{align}
where $\alpha$ denotes the fraction of active pixels in a patch on average.

It may be argued that the NVS may be operated in a traditional mode while a noise filter block can generate an interrupt for the processor after filtering noise events. We show next that our approach is more memory efficient than this one. A commonly used filter for event-based output stream is a Nearest Neighbour filter (NN-filt) \cite{Padala2018} that stores timestamps (using $B_t$ bits per timestamp) for every incoming event and marks this event as a noisy event if it does not get temporal support in a $p\times p$ spatial neighbourhood. The number of computes per event is then equal to $p^2-1$ comparisons and counter increments followed by one $B_t$ bit memory write to update the timestamp. The memory requirement for the two approaches may now be compared as:
\begin{align}
\label{eq:memory_filter}
    C_{NN-filt} &= (2(p^2-1)+B_t)\times \overline{n} \notag\\ 
    M_{NN-filt} &= B_t\times A\times B  
\end{align}
where $\overline{n}$ is the average number of events per frame
Note that $\overline{n}=\beta \times \alpha \times A\times B$ ($\beta\geq 1$) where $\beta$ represents average number of times an active pixel fires in $t_F$. Since the objects generally take up less than $10\%$ of the image, we have a conservative estimate of $C_{EBBI}=125.2$ kops/frame while $C_{NN-filt}\approx 276.4$ kops/frame. For the memory requirement, with typical values of $B_t=16$, our proposed method provides $8X$ memory savings. For the DAVIS sensor used, the reduced memory requirement of our proposed EBBI is only $10.8$ kB.

\subsection{Region Proposal}
\label{sec:rpn}
A region proposal network (RPN) is the first block in a tracking pipeline \cite{redmon2016you}. In our application of traffic monitoring using a stationary camera, NVS offers the advantage of almost perfect foreground background separation inherently since the pixels only respond to changes in contrast \cite{davis}. Thus background pixels will generate little or no events while moving objects will generate a significantly larger number of events. This allows us to locate valid regions without having to perform costly CNN operations \cite{rpn1}. 
\begin{figure}
    \centering
    \includegraphics[width=0.47\textwidth]{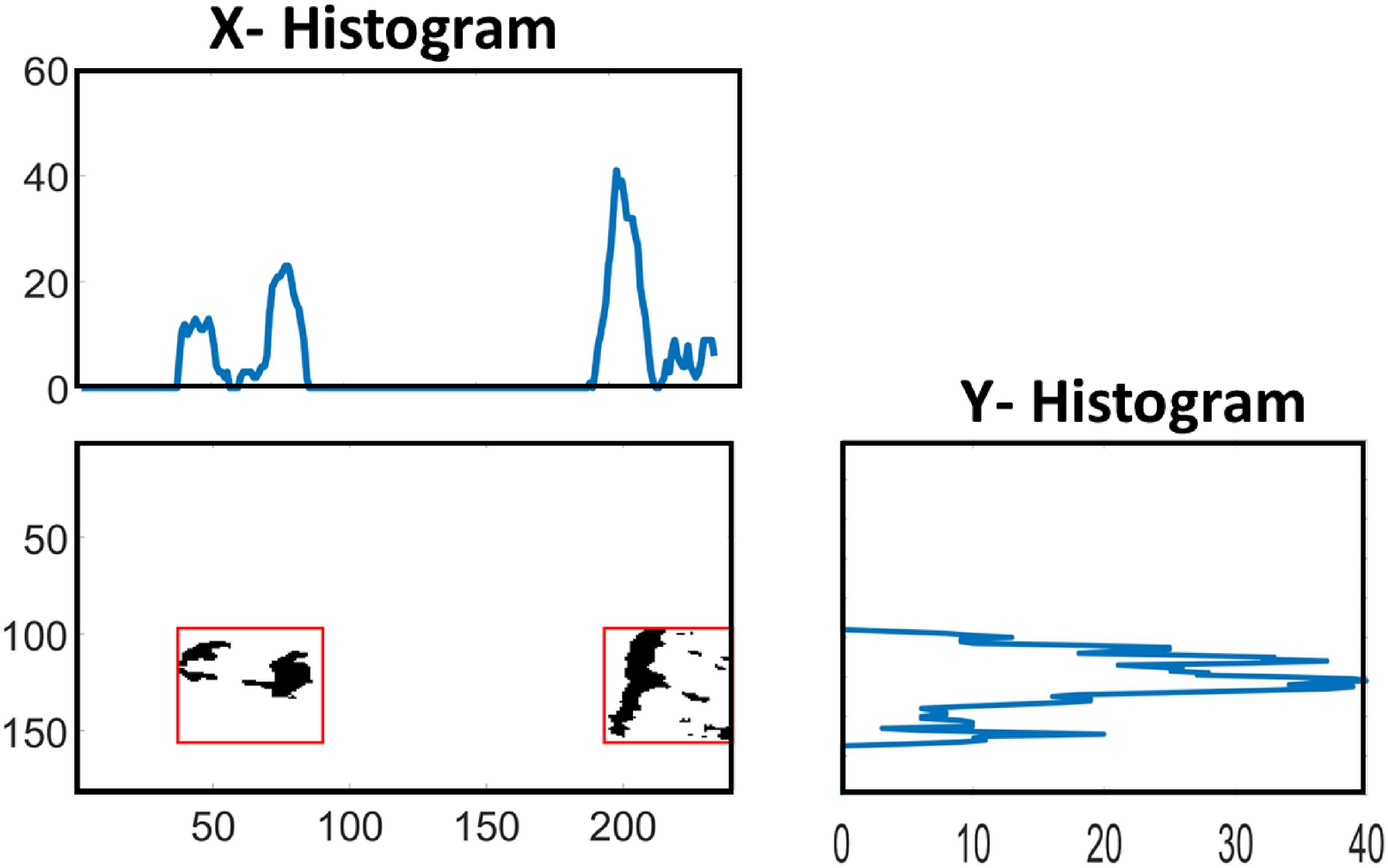}
    \caption{A sample EBBI with corresponding X and Y histogram based region proposals.}
    \label{fig:rpn}
    \vspace{-0.5cm}
\end{figure}

A traditional approach to detect regions in our case would have been to perform connected components analysis (CCA) on binary images using morphological operators \cite{Gonzalez:2006:DIP:1076432}. While this is still less costly than CNN operations, we propose a further simplified approach in our case by exploiting the fact that our application only requires a side view of the ongoing traffic. Instead of doing CCA on  a 2-D image, we create X and Y histograms ($H_X$ and $H_Y$) for the image and find regions in these two 1-D signals by finding consecutive entries that are higher than a threshold (Figure \ref{fig:rpn}). The RPN and the following tracker both operate on these 1-D data structures reducing the computational burden. The actual 2D region is obtained by finding intersections of the X and Y regions (Figure \ref{fig:rpn}). Histogram based location of objects for NVS has been earlier proposed \cite{Afshar2017}, but the authors only had one moving object without any realistic occlusion and  did not implement a full tracker on these region proposals.

To further reduce the computation and memory requirement, we create the histograms from a scaled image, $I^{s_1,s_2}$ downsampled from the original one ($I$) by factors $s_1$ and $s_2$ in X and Y directions respectively. Mathematically we can write the scaled image as:
\begin{align}
\label{eq:image_scale}
    I^{s_1,s_2}(i,j)=\sum_{n=0}^{s_2-1}\sum_{m=0}^{s_1-1}I(is_1+m,js_2+n)\notag\\
    i<\left\lfloor A/s_1\right\rfloor, j<\left\lfloor B/s_2\right\rfloor
\end{align}
where $I(i,j)\epsilon \{0,1\}$. Based on this, the histograms are defined as:
\begin{align}
\label{eq:image_scale}
    H_X^{s_1}(i)=\sum_j I^{s_1,s_2}(i,j)\notag\\
    H_Y^{s_2}(j)=\sum_i I^{s_1,s_2}(i,j)
\end{align}
X and Y regions are then found from $H_X^{s_1}$ and $H_Y^{s_2}$ by finding contiguous elements that are higher than a threshold (set to $1$ in this case). This is acceptable since we need a coarse location for the objects which will be smoothed by the tracker. In fact, this helps in overcoming \emph{fragmentation} of the object into smaller parts. As an example, the car in Fig. \ref{fig:rpn} displays two peaks in $H_X$ and would normally generate two separate regions. But in the low resolution histogram $H^{s_1}_X$, these mini regions get merged to create one region albeit with a slightly larger size than desired. One issue with this approach is that if there are multiple regions in both X and Y directions, false regions may be proposed by considering all overlaps between the two. In such cases, a check needs to be done in the original image to see if there are any valid pixels in that region. A better solution in that case is to perform the 2-D CCA, a task which we leave for future generalization of this approach. The total number of computes and memory requirement may now be summarized as follows:
\begin{align}
\label{eq:memory_rpn}
    C_{RPN} &=  A\times B + 2\frac{A\times B}{s_1s_2}\notag\\
    M_{RPN} &= \frac{A\times B}{s_1s_2} \{\lceil log_2(s_1s_2)\rceil\}+ \notag\\ 
    &(\frac{A}{s_1}\{\lceil log_2(B\times s_1) \rceil\} + \frac{B}{s_2}\{\lceil log_2(A\times s_2) \rceil\})
\end{align}
Here the first term for $C_{RPN}$ ($M_{RPN}$) denotes the computes (memory) needed for $I^{s_1,s_2}$ while the second term denotes the same for $H^{s_1}_X$ and $H^{s_2}_Y$. For our specific case, $s_1=6$ and $s_2=3$ were found to work well and in that case, $C_{RPN}=45.6$ kop/frame while $M_{RPN}\approx 1.6$ kB. Both of the equations are dominated by the first term.
In comparison to this, even the simplest CNN-based object detector like YOLO \cite{redmon2016you} would need GPUs for real-time performance (30 fps) with RAM usage in the order of Gigabytes ($> 1 GB$).
\vspace{-0.2cm}


\subsection{Overlap Based Tracker}
\label{sec:tracker}
In this paper we implement a multi-tracker system that can have up to $N_T=8$ trackers simultaneously active on the frame. Two main assumptions have driven the design of this simplified tracker:
\begin{itemize}
    \item $t_F$ is small enough such that there is significant overlap between objects from one frame to next.
    \item Distractors such as trees which create spurious events can be removed by a manually provided definition of region of exclusion (ROE). Static occlusion from posts etc can also be included in ROE.
\end{itemize}
A major issue faced by trackers for EBBI is object \emph{fragmentation} as shown in Fig. \ref{fig:rpn} and discussed in Section \ref{sec:rpn}. This happens because big vehicles like a bus have a lot of plane surface on their sides that do not generate much events.

A block diagrammatic representation of the major steps in the algorithm are shown in Fig. \ref{fig:flowchart}. The basic philosophy is to make predictions about the current position of a tracker from earlier frame and then correcting it based on current region proposals \cite{Lin2015}. Denoting the position vectors (bottom left corner co-ordinates (x,y), width (w) and height (h) of tracker box) for region proposals and trackers by $P_i$ and $T_i$ respectively ($1\leq i\leq 8$), we summarize the major steps below for every frame:
\begin{enumerate}
    \item Get predicted position $T_i^{pred}(x,y)$ of all valid trackers by adding $T_i(x,y)$ with corresponding X and Y velocities.
    \item Match $T_i^{pred}$ for each valid tracker $i$ with all available region proposals $P_j$. A match is found if overlapping area between the two is larger than a certain fraction of area of $T_i^{pred}$ or $P_j$--hence the name overlap based tracker (OT). 
    \item If a $P_j$ has no match and there are available free trackers, seed a new tracker $k$ with $T_k=P_j$.
    \item If a $T_i^{pred}$ matches single or multiple $P_j$, assign all $P_j$ to it and update $T_i$ and velocities as a weighted average between prediction and region proposal. Here, past history of tracker is used to remove \emph{fragmentation} in current region proposal if multiple $P_j$ had matched.
    \item If a $P_j$ matched multiple $T_i^{pred}$, we can have two cases--dynamic occlusion between two moving objects or the earlier region proposals for this object were fragmented leading to multiple trackers being assigned to it. Occlusion is detected if the predicted trajectory of those trackers for upto $n=2$ future time steps result in overlap. In that case, $T_i$ is updated entirely based on $T_i^{pred}$ and previous velocities are retained. Otherwise, the multiple $T_i^{pred}$ are merged into one tracker based on $P_j$ and corresponding velocity updated. The other trackers are freed up for future use.
\end{enumerate}

The memory requirement for this tracker is negligible ($<0.5$ kB) compared to the other modules and can be implemented in registers. The computation depends on which of the above cases is true. An average number of computes per frame can be obtained as follows:
\begin{align}
\label{eq:compute_tracker}
    C_{OT} =  134\overline{N_T}^2+\gamma_3N_3+\gamma_4N_4+\gamma_5N_5
\end{align}
where $\overline{N_T}$ denotes the average number of valid trackers, $\gamma_j$ and $N_j$ denote probability and number of computes for step $j$ in the tracker description above. The first term dominates the others due to low values of $\gamma_j$. For the recordings used, $\overline{N_T}\approx2$ resulting in $C_{OT}\approx564$.

A Kalman Filter (KF) tracking algorithm based on \cite{Lin2015} was used as comparison. The implementation follows a constant velocity motion model, hence contains a state vector of length $2$ ($X_{centroid}, Y_{centroid}$) for each track. Using \cite{Valade2017} to approximate its computational complexity, Eq. \ref{eq:Kfilter} shows the approximate computes for a Kalman filter with $\overline{N_T}=2$ tracks. 
\begin{align}
\label{eq:Kfilter}
    C_{KF} =& 4m^3 + 6m^2n + 4mn^2 + 4n^3 + 3n^2
\end{align}

Where $n$ and $m$ are the state and measurement vector size, respectively. Therefore, for this implementation with $n=2\times{\overline{N_T}}$ and $m=2\times{\overline{N_T}}$, $C_{KF} = 1200$. 
Memory requirement of the KF is $\approx 1.1$ kB which is also much smaller compared to the earlier blocks in the processing pipeline.

As an example of an event based tracker to be used in a fully event based pipeline after NN-filt, we chose \cite{Delbruck2013}. For the event based mean shift (EBMS), average number of computes per frame ($C_{EBMS}$) and memory requirement in bits ($M_{EBMS}$) can be given as:
\begin{align}
\label{eq:EBMS}
    C_{EBMS} =& \overline{N_{F}} [9~\overline{CL}^2+{(169+16~\gamma_{merge})}~\overline{CL} +11]\notag\\
    M_{EBMS} =& 408 CL_{max}+ 56
\end{align}

Where $\overline{N_F}$ is the average number of events per frame at the output of NN-filter, $\overline{CL}$ is the average number of active clusters at any given time ($\approx \overline{N_T}$), $\gamma_{merge}$ is the probability of two clusters merging and $CL_{max}$ is the maximum number of potential clusters.For these calculations we have assumed that past 10 positions of a cluster is used to calculate the current velocity of the cluster using least square regression. For our experiments we have used $CL_{max}=8$ and  for our dataset, $\overline{CL}\approx 2$, $\gamma_{merge}\approx0.1$ and $\overline{N_F}\approx 650$. So, EBMS requires 252 kops per frame which is $\approx 500X$ higher than EBBIOT. The total memory required for EBMS is $3.32$ kB which is again negligible compared to the earlier processing blocks for noise filtering.

\section{Results}

\subsection{Datasets}

Address Event Representation (AER) based event data is acquired from a DAVIS setup at a traffic junction. This scenario captures moving entities in the scene and the typical objects in the scene include humans, bikes, cars, vans, trucks and buses. Two sample recordings of varying duration are obtained at different lens settings and the comprehensive details of those two recordings used in the paper are presented in table \ref{Table1:Dataset}. The sizes of various moving objects vary by an order of magnitude in any given scene and their velocities also range over a wide range (sub-pixel to 5-6 pixels/frame). These recordings were manually annotated to generate the Ground Truth tracker annotations of these objects in the scene.

\begin{table}[h]
\vspace{-0.2cm}
\caption{Dataset Details}
\centering
\begin{tabular}{ |p{1cm}|p{1cm}|p{1cm}|p{1.75cm}|p{1.2cm} | }
 \hline
 Location & Lens (mm) & Duration (s) & Num Events\\
 \hline
 ENG    &  12   & 2998.4    &   107.5M \\
 LT4    &  6    & 999.5     &   12.5M\\
 \hline
\end{tabular}
\label{Table1:Dataset}
\vspace{-0.5cm}
\end{table}

\subsection{Tracker Evaluation Metrics}

To assess the performance of the tracker, we examine how closely the tracks generated by the proposed tracker match with the ground truth tracks. The first step in this evaluation process involves obtaining the boxes encapsulating the objects in the scene from both Ground Truth and the proposed tracker annotations at multiple instances of time (with a fixed time interval) in the entire duration of the recording. For each instance, if the area of a ground truth box enclosing an object in the scene is $A_{GroundTruth}$, the area of a tracker box enclosing an object is $A_{ProposedTracker}$, the area of intersection of these two boxes is denoted as $A_{Intersection}$ and the area of the union of these two boxes is denoted as $A_{Union}$. We adopted Intersection over Union (IoU) typically used in computer vision community (as defined in \ref{eq:IoU}) to measure the effectiveness of our tracker algorithm.
\begin{equation}
\label{eq:IoU}
    IoU=\frac{A_{Intersection}}{A_{Union}}
\end{equation}


 The proposed tracker box is assumed to be a correct region proposal only if the IoU of that box is greater than a threshold. 

A certain parametric threshold is defined based on IoU (e.g. IoU 0.5) to determine if a proposed box is a correct region proposal. Proposal boxes with IoU values larger than that threshold value is considered correct region detection (true positive box). Then, the performance of the tracker is evaluated on precision (true positive boxes/total proposal boxes) and recall(true positive boxes/total ground truth boxes) calculated over all the frames of the video. 
\subsection{Performance and Resource Requirements}
Finally, for a fair comparison of the EBBIOT algorithm with EBMS and Kalman Filter (KF) we compare the weighted average of precision and recall across multiple recordings where the weights correspond to the number of ground truth tracks present in a given recording. The results are shown in fig. \ref{fig:algo_comp}.

\begin{figure}[!tb]
        \centering
         \includegraphics[width=0.48\textwidth]{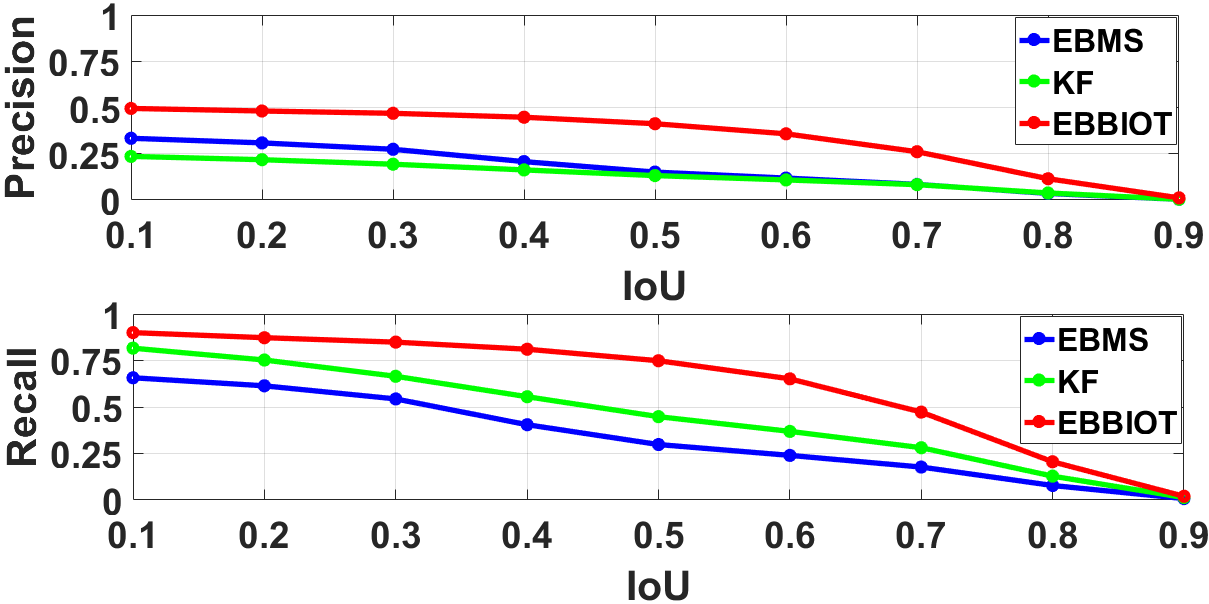}

    
    \caption{Comparison of EBMS, KF and EBBIOT in terms of precision and recall for different IoU thresholds. EBBIOT outperforms others and shows more stable precision and recall values for varying thresholds.}
    \label{fig:algo_comp}
\end{figure}
We also calculated total computes per frame and total memory required for KF and EBMS relative to EBBIOT (fig. \ref{fig:resource_comp}). For EBBIOT and KF total memory and computes are calculated considering memory and computes required for generating EBBI, RPN and tracker while for EBMS we consider memory and computes of NN-filt and EBMS tracker.
\begin{figure}[!htb]
        \centering
         \includegraphics[width=0.48\textwidth]{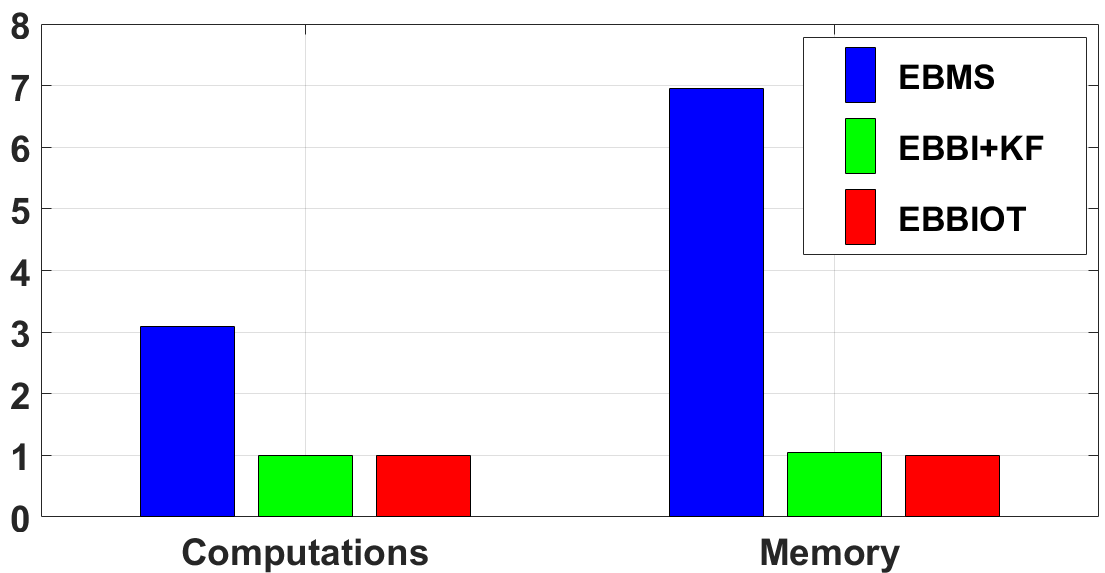}    
    \caption{Comparison of EBMS and EBBI+KF with EBBIOT in terms of total computations per frame and memory requirement showing significantly less resource requirement for EBBIOT. }
    \label{fig:resource_comp}
    \vspace{-0.5cm}
\end{figure}
\section{Conclusion}
\vspace{-0.1cm}
In this paper, we presented EBBIOT -- a novel paradigm for object tracking using stationary neuromorphic vision sensors in low-power sensor nodes for IoVT applications and demonstrated reliable tracking performance compared to both event-based tracking methodologies and traditional frame-based tracking frameworks. In particular, the mixed approach of creating event-based binary images resulted in $\approx 7X$ reduced memory requirement and $3X$ less computes over conventional event-based approaches. On the other hand, region proposals based on event density resulted in $>1000X$ less memory and computes compared to frame based approaches. Future work will change the RPN to a a general connected component approach \cite{Gonzalez:2006:DIP:1076432} instead of relying on side views. Further, we have not tracked slow and small objects like humans--this can be done by a two time scale approach where a second frame is generated with longer exposure times to capture activity of humans.



\bibliographystyle{IEEEtran}

\bibliography{RishBib}


  

%

\end{NoHyper}
\end{document}